\documentclass[conference, a4paper]{IEEEtran}
\IEEEoverridecommandlockouts
\usepackage{cite}
\usepackage{amsmath,amssymb,amsfonts}
\usepackage{algorithmic}
\usepackage{graphicx}
\usepackage{color}
\usepackage{textcomp}
\usepackage{subfig}
\usepackage{multirow}

\def\vector#1{\mbox{\boldmath $#1$}}
\def\BibTeX{{\rm B\kern-.05em{\sc i\kern-.025em b}\kern-.08em
    T\kern-.1667em\lower.7ex\hbox{E}\kern-.125emX}}
\begin{document}

\title{A Detection Method of Temporally Operated Videos Using Robust Hashing}

\author{\IEEEauthorblockN{Shoko Niwa}
\IEEEauthorblockA{\textit{Tokyo Metropolitan University}\\
Tokyo, Japan \\
niwa-shoko@ed.tmu.ac.jp}
\and
\IEEEauthorblockN{Miki Tanaka}
\IEEEauthorblockA{\textit{Tokyo Metropolitan University}\\
Tokyo, Japan \\
tanaka-miki@ed.tmu.ac.jp}
\and
\IEEEauthorblockN{Hitoshi Kiya}
\IEEEauthorblockA{\textit{Tokyo Metropolitan University}\\
Tokyo, Japan \\
kiya@tmu.ac.jp}
}

\maketitle

\begin{abstract}
SNS providers are known to carry out the recompression and resizing of uploaded videos/images, but most conventional methods for detecting tampered videos/images are not robust enough against such operations. In addition, videos are temporally operated such as the insertion of new frames and the permutation of frames, of which operations are difficult to be detected by using conventional methods. Accordingly, in this paper, we propose a novel method with a robust hashing algorithm for detecting temporally operated videos even when applying resizing and compression to the videos.
\end{abstract}


\section{Introduction}
Recent rapid advances in image manipulation tools and deep image synthesis techniques have made generating fake videos easy. In addition, with the spread of SNS, the existence of fake videos has become a major threat to the credibility of the international community. Accordingly, detecting tampered videos/images has become an urgent issue \cite{verdoliva2020media}. 

In particular, videos can easily be manipulated to produce fake videos by using the operations among frames such as the insertion, deletion, and permutation of video frames, called temporal operation. However, most methods for detecting forgery are not useful for such temporal operations. In addition, they are not robust enough against various types of content-preserving transforms without malice, such as resizing and compression. Since most SNSs are known to carry out such operations, conventional methods are efficient in such cloud environments~\cite{chuman2019image, chuman2017image}. To overcome the issue, various methods robust against the operations have been studied for still images~\cite{tanaka2021detection, iida2020privacy, arnia2006fast}. Accordingly, we propose a novel method for robustly detecting temporally operated videos using a robust hashing algorithm. The proposed method allows us not only to detect temporally operated videos with a high accuracy but to also reduce the amount of hash values by synthesizing an ``extended frame'' from multiple frames.

\section{Related Works}
\subsection{Robust Hashing}
Most hashing methods such as secure hash algorithms (SHA) generally output significantly different hash values from slightly different input data sets. In contrast, robust hashing methods are designed to output similar hash values from similar input data sets. Accordingly, hash values are robust against input data including distortion caused with compression and image resizing, so robust hashing has been used as a method for image retrieval. 
Hash values for fake-image detection are required to be robust enough against a number of types of image operation such as image compression and resizing since such operation does not convert the content of images, although the quality of the images is reduced. Therefore, we focus on using a robust hash method that aims to robustly retrieve images similar to query images. In contrast, hash values generated by using a robust hash method have to be sensitive to the influence of manipulation used for generating tampered images such as copy-move and GANs.
Under these requirements, various robust hashing methods~\cite{li2015robust,kozat2004robust,gong2012iterative,venkatesan2000robust, iida2019robust, itagaki2021robust, du2020image} have been compared in terms of sensitivity and robustness in which Li~\textit{et al}.'s method~\cite{li2015robust} was demonstrated to have a suitable performance for tampered image detection. 

In this paper, we also use Li~\textit{et al}.'s robust hashing method~\cite{li2015robust}, which was confirmed to have high performance for fake-image detection in~\cite{tanaka2021detection}. The method can capture both spatial and chromatic features by using quaternions. The method is carried out as follows:



\begin{enumerate}
  \item Apply a Gaussian filter with a kernel size of $5 \times 5$ and a standard deviation of 1 to the image and then resize it to $128 \times 128$ pixels.
  \item Extract spatial and chromatic features from the preprocessed image using quaternions.
  \item Select some features from them using a feature selection algorithm.
 \item Generate a binary hash value of 120 bits in length.
\end{enumerate}

\subsection{Tampering video}
\begin{figure}[tb]
\centering
\subfloat[Frame insertion]{\includegraphics[keepaspectratio,width=3.5cm]{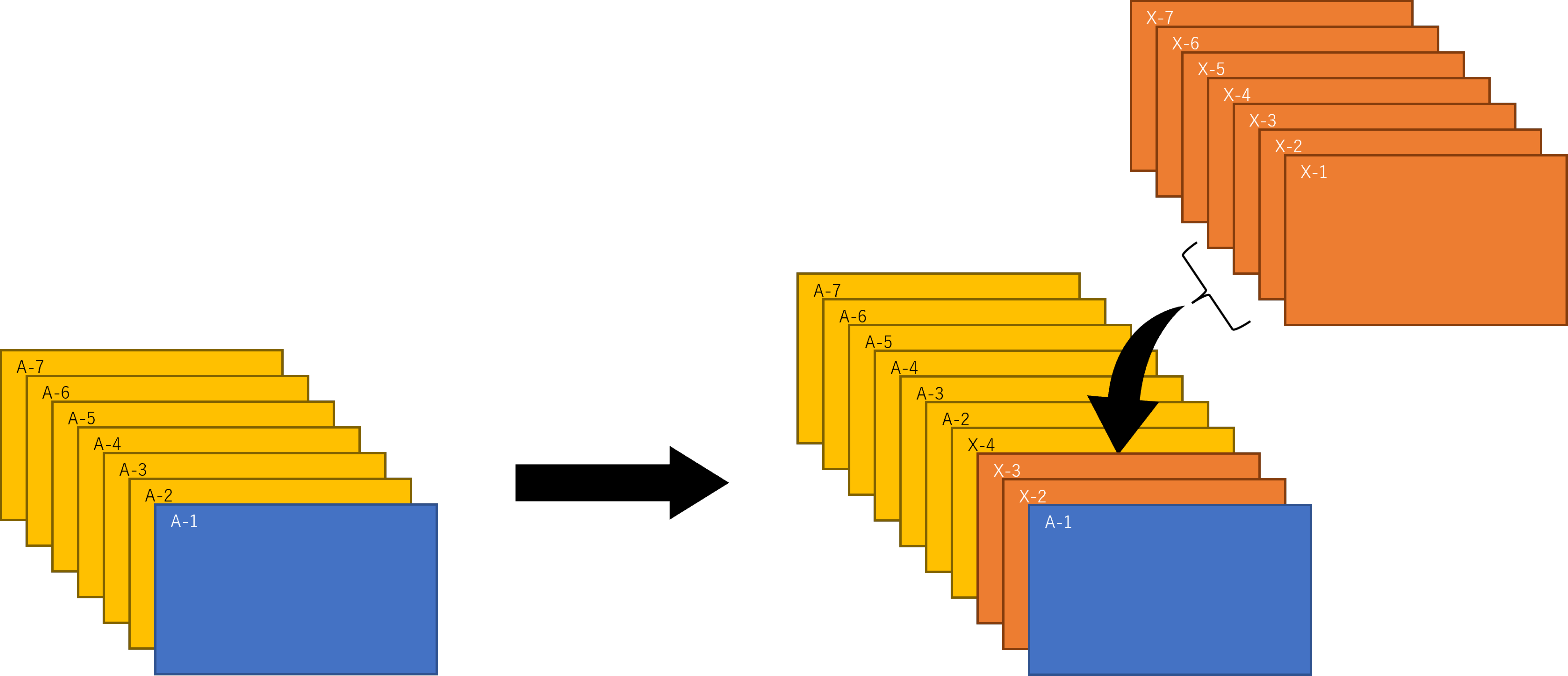}\label{interframe1}}
\hfil
\subfloat[Frame deletion]{\includegraphics[keepaspectratio,width=3.5cm]{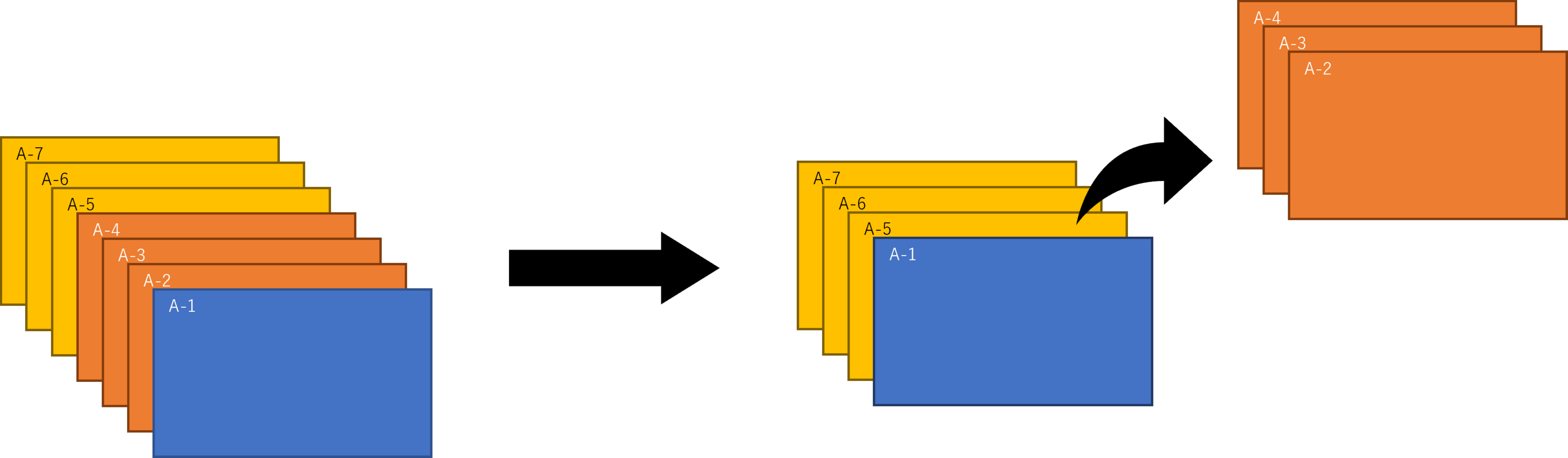}\label{interframe2}}\\
\quad
\subfloat[Frame reordering]{\includegraphics[keepaspectratio,width=2.8cm]{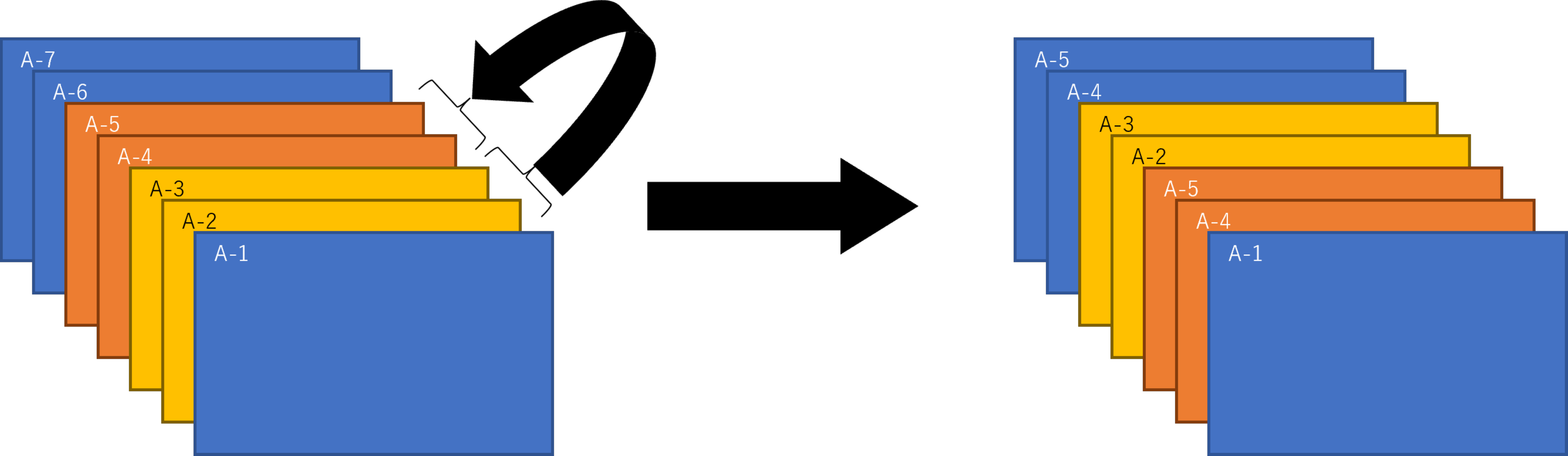}\label{interframe3}}
\hfil
\subfloat[Frame replace]{\includegraphics[keepaspectratio,width=3.8cm]{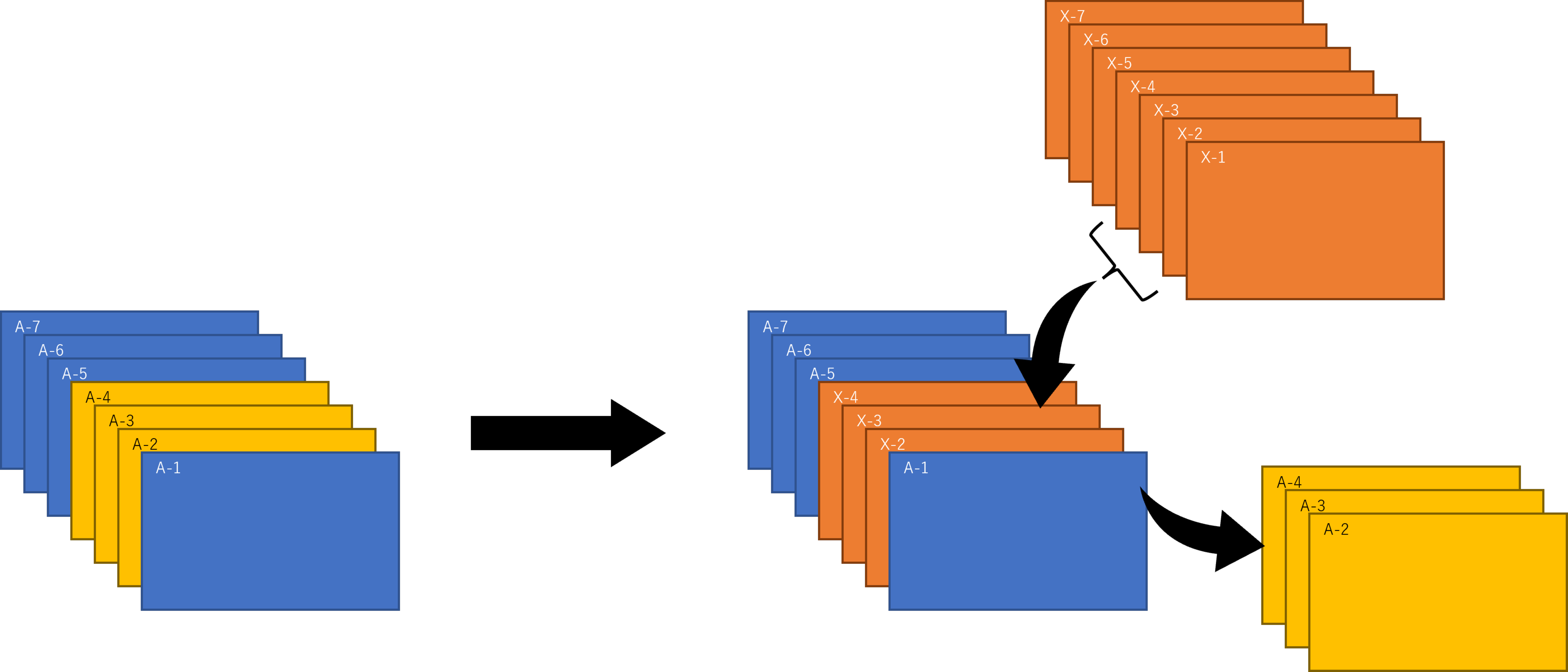}\label{interframe4}}

\caption{Example of inter-frame tampering}
\label{interframe}
\end{figure}

Recent rapid advances in image manipulation tools and deep image synthesis techniques have made tampering videos easy~\cite{nirkin2019fsgan, elharrouss2020image}. Video tampering is classified into two types: intra-frame tampering and inter-frame tampering~\cite{milani2012overview}. 
Intra-frame tampering, also called spatial tampering, is an attack on each frame. Examples include adding (copy-move, splicing) and deleting (inpainting) objects. Adding objects is a method for cutting some objects from other frames and pasting them into the original frame. Deleting objects hides some objects in the original frame with a background color.
In contrast, inter-frame tampering, also called temporal tampering, is to manipulate the relationships among frames. Examples include inserting, deleting, rearranging, and replacing frames as shown in Fig.~\ref{interframe}. 

Frame insertion (Figure~\ref{interframe1}) is a technique that inserts some frames from another video frames into original ones. Frame deletion (Figure~\ref{interframe2}) is to delete some frames from the original video. Frame reordering (Figure~\ref{interframe3}) is to change the order of some or all of the frames in the original video. Frame replace (Figure~\ref{interframe4}) is to replace some frames in the original video with some frames in another video.
In this paper, we focus on detecting intra-frame tampering, which is difficult for other conventional methods.

\section{Proposed Method}\label{tejun}
\begin{figure}[htb]
    \centering
    \includegraphics[keepaspectratio,width=8.8cm]{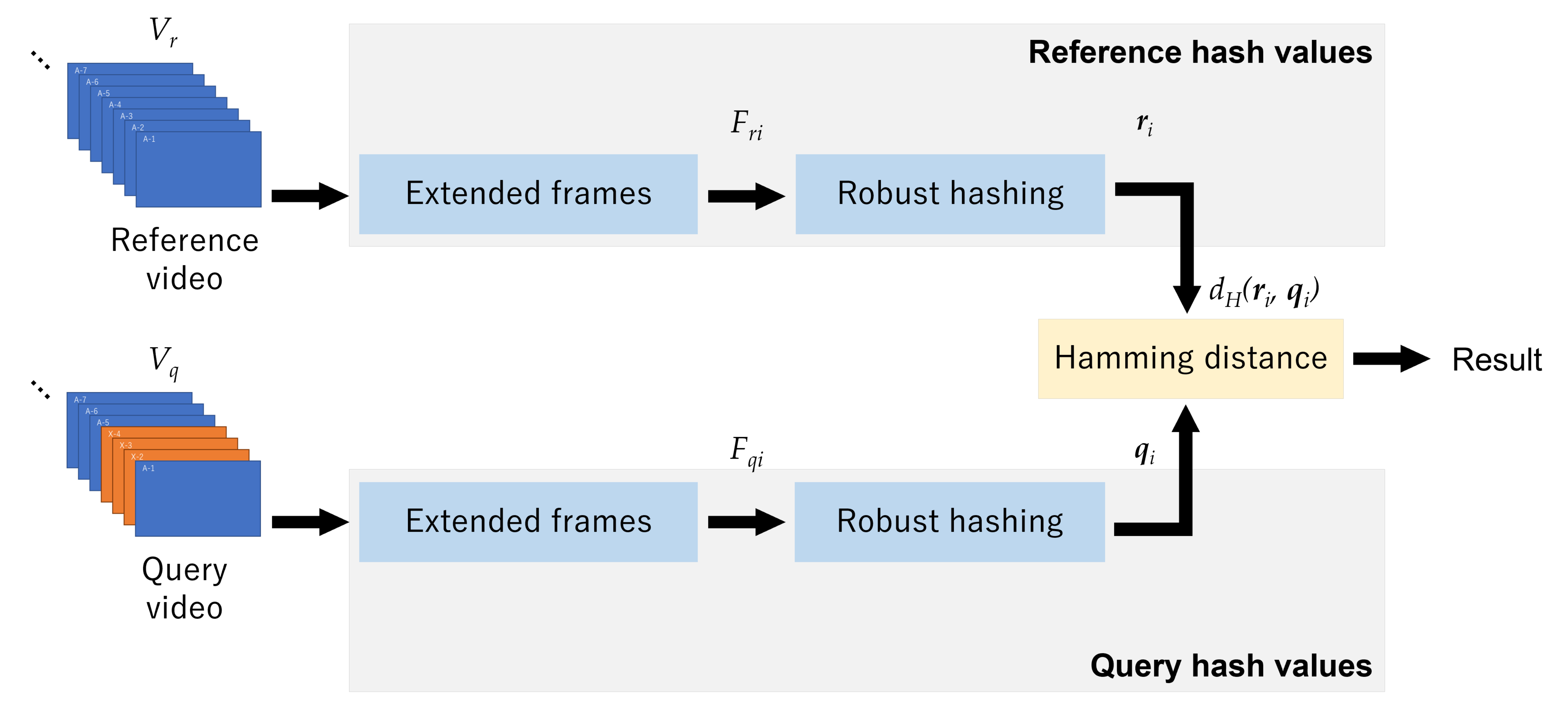}
    \caption{Overview of proposed method}
    \label{gaiyou1}
\end{figure}

An overview of the proposed method is shown in Fig.~\ref{gaiyou1}.
A reference video $V_r = \{ f_{r1},f_{r2},\ldots,f_{rl} \}$ and a query video $V_q = \{ f_{q1},f_{q2},\ldots,f_{ql} \}$ are prepared, where $V_q$ and $V_r$ consist of $l$ frames respectively, and $f_{ri}$ and $f_{qi}$ indicate frames. 
Extended frames $F_{ri}, F_{qi}, i = 1, 2,\ldots$ are defined by using the frames, respectively, where each extended frame is produced by using $n\times n$ frames and then a hash value is computed from every extended frame by using Li~\textit{et al}.'s hash method~\cite{li2015robust}. Li~\textit{et al}.'s hash method generates a hash value with $J=120$ bits from an extended frame, so the Hamming distance between two hash values is computed as 

\begin{equation}
 \label{hamming}
d_H(\vector{r}_i, \vector{q}_i) \triangleq \sum_{k=1}^{J} \delta(r_{ik}, q_{ik}),
\end{equation}
where,
\begin{equation*}
\delta(r_{ik}, q_{ik}) =
\begin{cases}
1 & (r_{ik} \neq  q_{ik})\\
0 & (r_{ik}=q_{ik})
\end{cases}.
\end{equation*}
$\vector{r}_i=\{ r_{i1},r_{i2},\ldots,r_{ik},\ldots,r_{iJ}\}$ and $\vector{q}_i=\{ q_{i1}, q_{i2},\ldots,q_{ik},\\\ldots, q_{iJ}\}$, $r_{ik}, q_{ik}\in\{0,1\}$ are hash values computed from the $i$-th extended frames of $F_{ri}$ and $F_{qi}$.
In this method, $F_{qi}$ is judged as an operated frame if $d_H(\vector{r}_i, \vector{q}_i)$ is greater than or equal to the threshold $d$, as follows:

\begin{equation}
\label{eq:d}
F_{qi}=
\begin{cases}
\text{operated}& \text{if $d_H(\vector{r}_i, \vector{q}_i) \geq d$}\\
\text{non-operated}& \text{if $d_H(\vector{r}_i, \vector{q}_i) < d$}
\end{cases}.
\end{equation}

\subsection{Generation of extended frames}
\begin{figure}[tb]
    \centering
    \includegraphics[keepaspectratio,width=8.8cm]{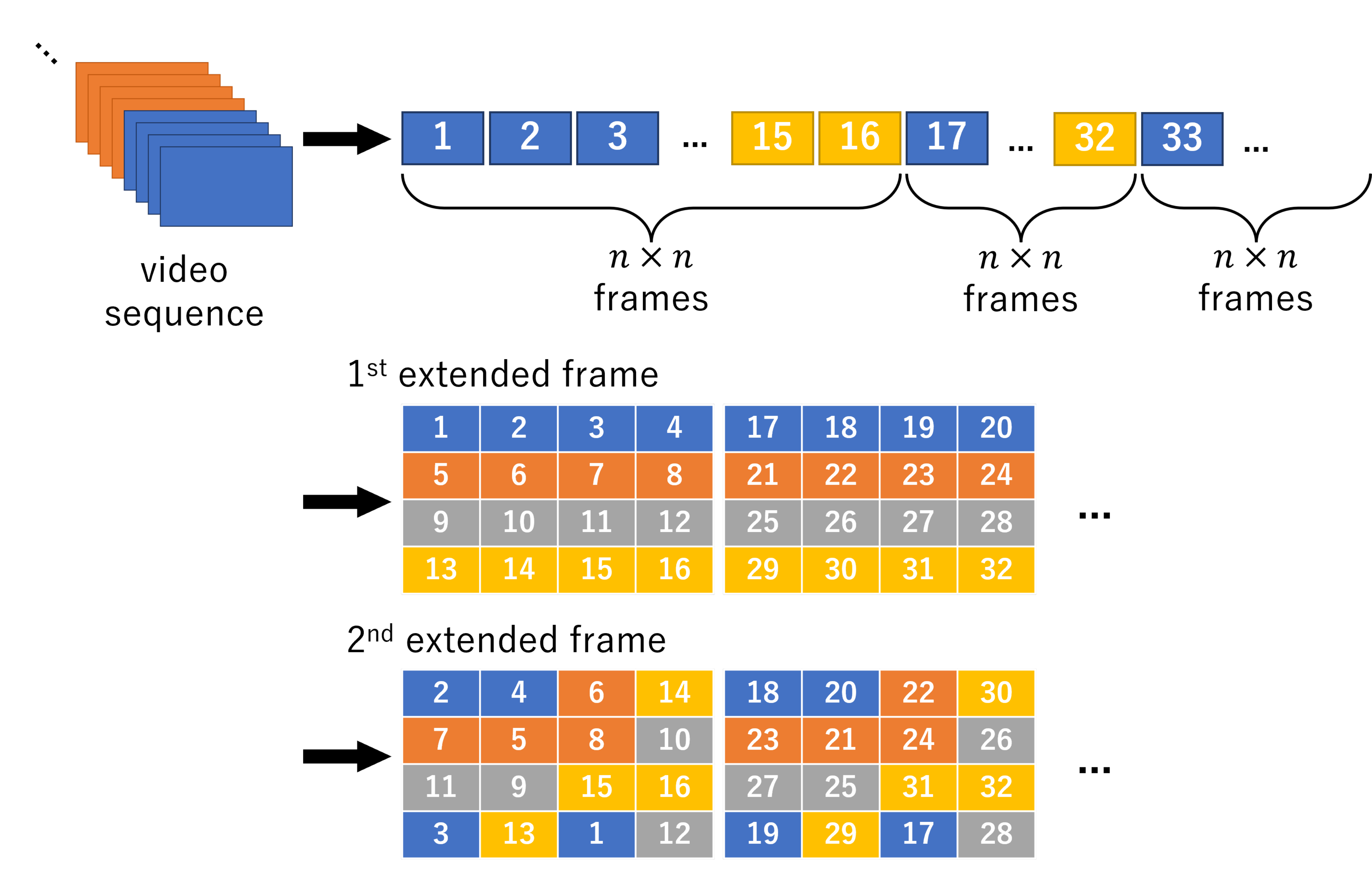}
    \caption{Generating extended frame $(n=4)$}
    \label{extend}
\end{figure}

Extended frames are generated from a reference video and a query video, respectively, as follows.
\begin{enumerate}[\IEEEsetlabelwidth{A-}]
   \item[(A-1)] A video sequence is divided into frame-blocks, each with $n\times n$ frames (see Fig.~\ref{extend}).
    
   \item[(A-2)] The first extended frame is defined with each frame-block in the order of the frame number (see Fig.~\ref{extend}). 
   
   \item[(A-3)] The second extended frame is defined with each frame-block such that frames near the four corners of the first extended frame are moved to the center of the extended frame.
   
\end{enumerate}

A property of Li~\textit{et al}.'s robust hashing method gives the reason that a few frames are permutated in (A-3). The hashing method uses the quaternion polar cosine transform\cite{li2015robust}, so the four corners of an image are not sensitive.



To investigate the necessity of extended frames, we experimented with two videos: a reference video $V_r$ consisting of $n\times n$ white frames, which was used for generating one extended frame $F_r$, and a query video $V_q$ consisting of $n\times n-1$ white frames and one black frame, which was used for generating one extended frame $F_q$, where the black frame in $F_q$ corresponded to a tampered frame.

\begin{figure}[tb]
\centering
\subfloat[With 1st extended frame]{\includegraphics[keepaspectratio,height=3.35cm]{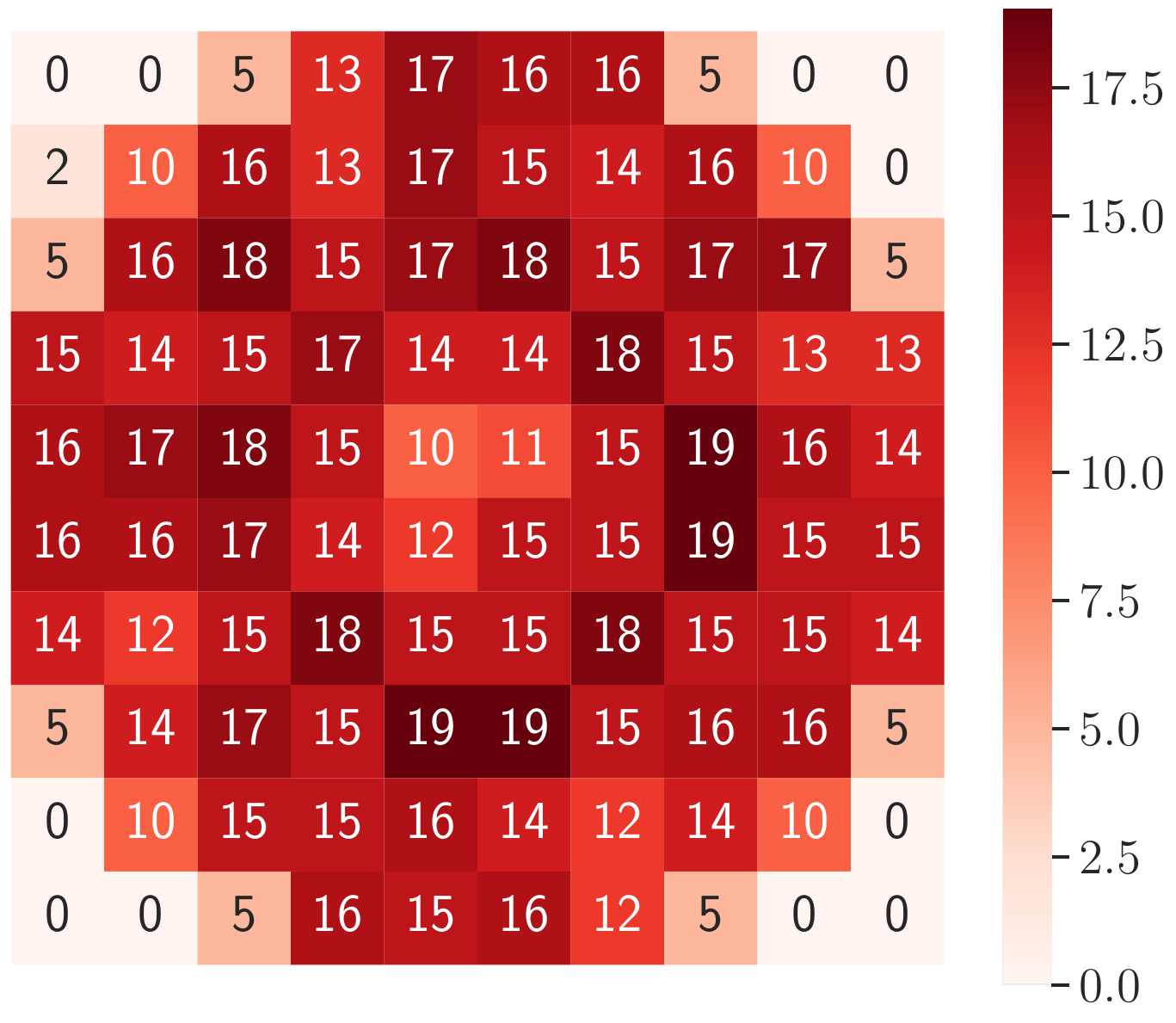}\label{notikan}}
\hfil
\subfloat[With both 1st and 2nd extended frames]{\includegraphics[keepaspectratio, height=3.35cm]{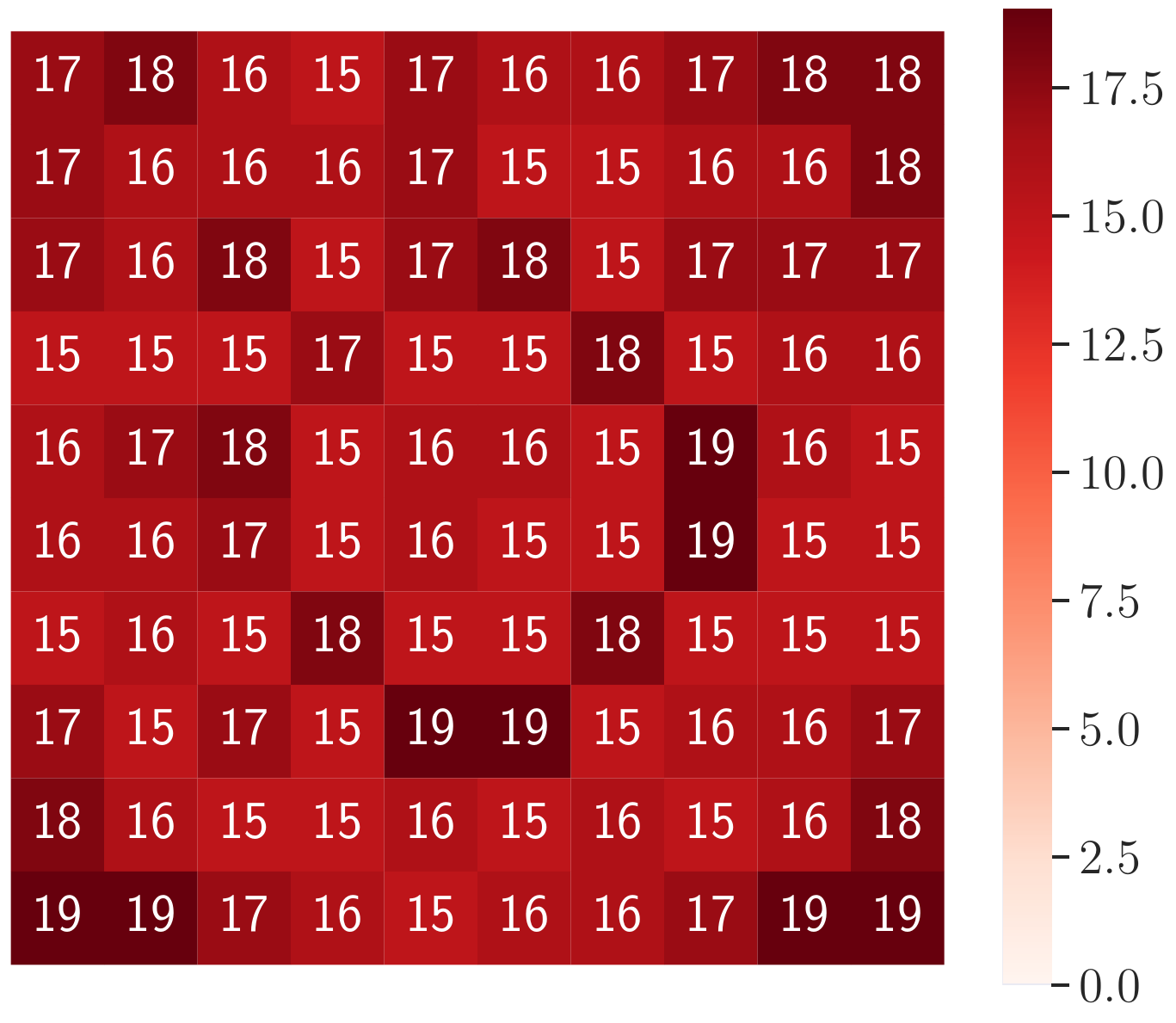}\label{MAX}}
\caption{Relationship between the position of the tampered frame and the Hamming distance $(n = 10)$}
\label{fakeframe-hamming}
\end{figure}

Figure~\ref{notikan} shows the relationship between the position of the black frame and the Hamming distance between $F_{r}$ and $F_{q}$.
If the black frame is at the $x$-th position, the Hamming distance between $F_{r}$ and $F_{q}$ is shown to be at the element~$(a,b)$, $(0 \leq a,b\leq n, x=(a-1)\times n+b)$ in Fig.~\ref{fakeframe-hamming}.
As you can see from Fig.~\ref{fakeframe-hamming}, in case of using Li~\textit{et al}.'s robust hashing method, the Hamming distance between $F_{r}$ and $F_{q}$ becomes almost zero value when the black frame is near the four corners of $F_{q}$.
Therefore, to avoid putting tampered frames near the four corners of the image, in this paper, we propose to prepare two types of extended frames: those ordered in the usual order. and those ordered in the replaced order, as shown in Fig.~\ref{notikan}. This technique is expected to make it easier to detect tampering even if the tampered frames are located near the four corners of extended frames (see Fig.~\ref{MAX}).

\subsection{Detection procedure}
The procedure of the proposed method is summarized as below.
\begin{enumerate}[\IEEEsetlabelwidth{B-}]
\item[(B-1)] Two types of extended frames are generated from a reference video.
\item[(B-2)] 
A hash value is computed from every extended frame by using Li~\textit{et al}.'s robust hashing method, and the values are stored.

\item[(B-3)] Two types of extended frames are generated from a query video, and a hash value is computed from every extended frame by using Li~\textit{et al}.'s robust hashing method as well.

\item[(B-4)]Hash values are compared between an extended reference frame and the corresponding extended query frame to detect whether the query video is operated by Eqs.~\eqref{hamming} and \eqref{eq:d}.


\end{enumerate}


\section{Exreriment}
\subsection{Experimental setup}
To confirm the effectiveness of the proposed method, six datasets were prepared by ourselves as shown in Table~\ref{dataset}. Each dataset consists of frames of size $1280\times 720$ pixels, with a total of 3540 frames, 540 of which are tampered frames. 
Three videos were downloaded from a web page, and dataset 0-0 was prepared from the three videos as a reference dataset. Dataset 0-0 was uploaded to Twitter and Instagram, and then dataset 0-1 and dataset 0-2 were produced by downloading dataset 0-0 from Twitter and Instagram, respectively. Thus, both datasets were not equal to dataset 0-0 due to the influence of recompressing and resizing. 
Dataset 1-0 was produced by applying a number of frames in dataset 0-0 to temporal manipulations, i.e. inter-frame tamper operations such as frame deletion, reordering, and replacement. dataset 1-1 was produced by uploading dataset 1-0 to Twitter, and dataset 1-2 was produced by uploading dataset 1-0 to Instagram.

\begin{table}[tb]
  \caption{Video datasets used in experiments}
  \label{dataset}
  \centering
  \scalebox{0.88}{
  \begin{tabular}{llcccc}
    \hline
      Dataset & \,\, Information & Compression & Resize & Frame size\\
    \hline \hline
    Dataset 0-0 & \,\, Reference & - & - & $1280\times 720$\\
    Dataset 0-1 & \,\, Uploaded to Twitter & + & - & $1280\times 720$\\
    Dataset 0-2 & \,\, Uploaded to Instagram & + & + & $1152\times 648$\\
    \hline
    Dataset 1-0 &
    \begin{tabular}{l}
    Including\\[-0.7mm] operated frames
    \end{tabular} & - & - & $1280\times 720$\\
    Dataset 1-1 & \,\, Uploaded to Twitter & + & - & $1280\times 720$\\
    Dataset 1-2 & \,\, Uploaded to Instagram & + & + & $1152\times 648$\\
    \hline
  \end{tabular}}
  \begin{flushleft}
  ``+'' indicates that frames in that dataset were compressed/resized, while ``-'' indicates that any frames in that dataset were not compressed/resized.
  \end{flushleft}
\end{table}

\begin{table}[t]
\caption{Definition of indicators}
\label{dif}
\centering
\begin{tabular}{cc|cc}
 &  & \multicolumn{2}{c}{Predict label} \\ \cline{3-4} 
 &  & Positive & Negative \\ \hline
\multicolumn{1}{c|}{Actual} & Positive & $\mathit{TP}$ & $\mathit{FN}$ \\
\multicolumn{1}{c|}{label} & Negative & $\mathit{FP}$ & $\mathit{TN}$
\end{tabular}
\end{table}

\subsection{Result}
The proposed method was evaluated by using two evaluation indexes: accuracy (Acc) and average precision (AP). Both Acc and AP are in the range of $[0,1]$, and a higher value indicates that tampered frames are detected more correctly. Acc and AP are given by
\begin{equation}
Acc = \frac{\mathit{TP} +\mathit{TN}}{\mathit{TP}+\mathit{FP}+\mathit{TN}+\mathit{FN}},
\end{equation}
\begin{equation}
AP = \sum_j (R_j - R_{j-1})P_j,
\end{equation}
where $\mathit{TP},\mathit{FP}, \mathit{TN}$ and $\mathit{FN}$ are as defined in Table~\ref{dif}, $P_j, R_j$ are precision and recall at the $j$-th threshold also given by $P_j = {\mathit{TP}_j}/(\mathit{TP}_j+\mathit{FP}_j)$, $R_j = {\mathit{TP}_j}/(\mathit{TP}_j+\mathit{FN}_j)$.

Experiment results are shown in Table~\ref{compare} where dataset 0-0 was used as a reference video, and two parameters $n$ and $d$ were experimentally decided as $n = 8$ and $d = 23$. From the results of dataset 0-1 and dataset 0-2, the method was confirmed to be robust enough against recompression and resizing operations. In addition, From the results of datasets 1-0, 1-1 and 1-2, it was verified to be effective in detecting temporally operated videos. In particular, when using two extend frames, the proposed method had a higher detection accuracy.

\begin{table}[tb]
  \caption{Comparison of tampering detection results \\for each extended frame$(n = 8, d = 23)$}
  \label{compare}
  \centering
  \begin{tabular}{ccccc}
    \hline
    Query&\multicolumn{2}{c}{With 1st extended frames}&\multicolumn{2}{c}{With both 1st and 2nd ones}\\
    \cline{2-5}
    dataset & Acc & AP & Acc & AP\\
    \hline \hline
    Dataset 0-1 & 1.0000 & not defined & 1.0000 & not defined\\
    Dataset 0-2 & 1.0000 & not defined & 1.0000 & not defined\\
    \hline
    Dataset 1-0 & 0.9825 & 1.0000 & 1.0000 & 1.0000\\
    Dataset 1-1 & 0.9825 & 1.0000 & 1.0000 & 1.0000\\
    Dataset 1-2 & 0.9825 & 0.9936 & 1.0000 & 1.0000\\
    \hline
  \end{tabular}
\end{table}

\section{Conclusion}
In this paper, we proposed a novel method with a robust hashing algorithm for detecting temporally operated videos where Li \textit{et al}.'s robust hashing method was used as a robust hashing algorithm. In addition, the method uses an approach called ``extended frames'' to detect tampered frames with high accuracy while referring to many frames at once. In an experiment, the method was demonstrated not only to be robust against image recompression and resizing but to also give a higher accuracy by using two extended frames.


\bibliographystyle{IEEEtran}
\bibliography{ref}

\begin{thebibliography}{10}
\providecommand{\url}[1]{#1}
\csname url@samestyle\endcsname
\providecommand{\newblock}{\relax}
\providecommand{\bibinfo}[2]{#2}
\providecommand{\BIBentrySTDinterwordspacing}{\spaceskip=0pt\relax}
\providecommand{\BIBentryALTinterwordstretchfactor}{4}
\providecommand{\BIBentryALTinterwordspacing}{\spaceskip=\fontdimen2\font plus
\BIBentryALTinterwordstretchfactor\fontdimen3\font minus
  \fontdimen4\font\relax}
\providecommand{\BIBforeignlanguage}[2]{{%
\expandafter\ifx\csname l@#1\endcsname\relax
\typeout{** WARNING: IEEEtran.bst: No hyphenation pattern has been}%
\typeout{** loaded for the language `#1'. Using the pattern for}%
\typeout{** the default language instead.}%
\else
\language=\csname l@#1\endcsname
\fi
#2}}
\providecommand{\BIBdecl}{\relax}
\BIBdecl

\bibitem{verdoliva2020media}
L.~Verdoliva, ``Media forensics and deepfakes: an overview,'' \emph{IEEE
  Journal of Selected Topics in Signal Processing}, vol.~14, no.~5, pp.
  910--932, 2020.

\bibitem{chuman2019image}
T.~CHUMAN, K.~IIDA, W.~SIRICHOTEDUMRONG, and H.~KIYA, ``Image manipulation
  specifications on social networking services for encryption-then-compression
  systems,'' \emph{IEICE Transactions on Information and Systems}, vol. E102.D,
  no.~1, pp. 11--18, 2019.

\bibitem{chuman2017image}
T.~Chuman, K.~Iida, and H.~Kiya, ``Image manipulation on social media for
  encryption-then-compression systems,'' in \emph{2017 Asia-Pacific Signal and
  Information Processing Association Annual Summit and Conference (APSIPA
  ASC)}.\hskip 1em plus 0.5em minus 0.4em\relax IEEE, 2017, pp. 858--863.

\bibitem{tanaka2021detection}
M.~Tanaka, S.~Shiota, and H.~Kiya, ``A detection method of operated fake-images
  using robust hashing,'' \emph{Journal of Imaging}, vol.~7, no.~8, p. 134,
  2021.

\bibitem{iida2020privacy}
K.~Iida and H.~Kiya, ``Privacy-preserving content-based image retrieval using
  compressible encrypted images,'' \emph{IEEE Access}, vol.~8, pp.
  200\,038--200\,050, 2020.

\bibitem{arnia2006fast}
F.~Arnia, I.~Iizuka, M.~Fujiyoshi, and H.~Kiya, ``Fast and robust
  identification methods for jpeg images with various compression ratios,'' in
  \emph{2006 IEEE International Conference on Acoustics Speech and Signal
  Processing Proceedings}, vol.~2, 2006, pp. II--II.

\bibitem{li2015robust}
Y.~N. Li, P.~Wang, and Y.~T. Su, ``Robust image hashing based on selective
  quaternion invariance,'' \emph{IEEE signal processing letters}, vol.~22,
  no.~12, pp. 2396--2400, 2015.

\bibitem{kozat2004robust}
S.~Kozat, R.~Venkatesan, and M.~Mihcak, ``Robust perceptual image hashing via
  matrix invariants,'' in \emph{International Conference on Image Processing},
  vol.~5, 2004, pp. 3443--3446.

\bibitem{gong2012iterative}
Y.~Gong, S.~Lazebnik, A.~Gordo, and F.~Perronnin, ``Iterative quantization: A
  procrustean approach to learning binary codes for large-scale image
  retrieval,'' \emph{IEEE transactions on pattern analysis and machine
  intelligence}, vol.~35, no.~12, pp. 2916--2929, 2012.

\bibitem{venkatesan2000robust}
R.~Venkatesan, S.-M. Koon, M.~H. Jakubowski, and P.~Moulin, ``Robust image
  hashing,'' in \emph{Proceedings 2000 International Conference on Image
  Processing (Cat. No. 00CH37101)}, vol.~3.\hskip 1em plus 0.5em minus
  0.4em\relax IEEE, 2000, pp. 664--666.

\bibitem{iida2019robust}
K.~IIDA and H.~KIYA, ``Robust image identification with dc coefficients for
  double-compressed jpeg images,'' \emph{IEICE Transactions on Information and
  Systems}, vol. E102.D, no.~1, pp. 2--10, 2019.

\bibitem{itagaki2021robust}
T.~Itagaki, Y.~Funabiki, and T.~Akishita, ``Robust image hashing for detecting
  small tampering using a hyperrectangular region,'' in \emph{2021 IEEE
  International Workshop on Information Forensics and Security (WIFS)}, 2021,
  pp. 1--6.

\bibitem{du2020image}
L.~Du, Z.~He, Y.~Wang, X.~Wang, and A.~T. Ho, ``An image hashing algorithm for
  authentication with multi-attack reference generation and adaptive
  thresholding,'' \emph{Algorithms}, vol.~13, no.~9, p. 227, 2020.

\bibitem{nirkin2019fsgan}
Y.~Nirkin, Y.~Keller, and T.~Hassner, ``Fsgan: Subject agnostic face swapping
  and reenactment,'' in \emph{Proceedings of the IEEE/CVF international
  conference on computer vision}, 2019, pp. 7184--7193.

\bibitem{elharrouss2020image}
O.~Elharrouss, N.~Almaadeed, S.~Al-Maadeed, and Y.~Akbari, ``Image inpainting:
  A review,'' \emph{Neural Processing Letters}, vol.~51, no.~2, pp. 2007--2028,
  2020.

\bibitem{milani2012overview}
S.~Milani, M.~Fontani, P.~Bestagini, M.~Barni, A.~Piva, M.~Tagliasacchi, and
  S.~Tubaro, ``An overview on video forensics,'' \emph{APSIPA Transactions on
  Signal and Information Processing}, vol.~1, p.~e2, 2012.

\end{thebibliography}

\end{document}